# DRIFT-FREE INDOOR NAVIGATION USING SIMULTANEOUS LOCALIZATION AND MAPPING OF THE AMBIENT HETEROGENEOUS MAGNETIC FIELD

J. C. K. Chow

Xsens Technologies, Pantheon 6a, 7521 PR Enschede, The Netherlands - jacky.chow@xsens.com

**ABSTRACT:**

In the absence of external reference position information (e.g. surveyed targets or Global Navigation Satellite Systems) Simultaneous Localization and Mapping (SLAM) has proven to be an effective method for indoor navigation. The positioning drift can be reduced with regular loop-closures and global relaxation as the backend, thus achieving a good balance between exploration and exploitation. Although vision-based systems like laser scanners are typically deployed for SLAM, these sensors are heavy, energy inefficient, and expensive, making them unattractive for wearables or smartphone applications. However, the concept of SLAM can be extended to non-optical systems such as magnetometers. Instead of matching features such as walls and furniture using some variation of the Iterative Closest Point algorithm, the local magnetic field can be matched to provide loop-closure and global trajectory updates in a Gaussian Process (GP) SLAM framework. With a MEMS-based inertial measurement unit providing a continuous trajectory, and the matching of locally distinct magnetic field maps, experimental results in this paper show that a drift-free navigation solution in an indoor environment with millimetre-level accuracy can be achieved. The GP-SLAM approach presented can be formulated as a maximum a posteriori estimation problem and it can naturally perform loop-detection, feature-to-feature distance minimization, global trajectory optimization, and magnetic field map estimation simultaneously. Spatially continuous features (i.e. smooth magnetic field signatures) are used instead of discrete feature correspondences (e.g. point-to-point) as in conventional vision-based SLAM. These position updates from the ambient magnetic field also provide enough information for calibrating the accelerometer bias and gyroscope bias in-use. The only restriction for this method is the need for magnetic disturbances (which is typically not an issue for indoor environments); however, no assumptions are required for the general motion of the sensor (e.g. static periods).

## 1. INTRODUCTION

With the VR/AR market predicted to be worth $108 billion by 2021 (Digi-Capital, 2017) one important technical challenge is indoor position and orientation tracking. AR in a shopping mall, classroom, or a volume too large for cost effective deployment of cameras can benefit from passive position tracking. Furthermore, head-tracking for AR/VR in harsh environments or confined spaces (e.g. aircraft cockpit) where it is difficult to maintain a camera or the field of view is too restricted, requires alternate sensors. MEMS inertial measurement units (IMUs) are getting smaller, cheaper, and more ubiquitous. For example, they are found in most smartphones, drones, and wearables these days. Although they can provide position information, their use is usually limited to orientation tracking. The reason is that small inertial errors can result in large positional errors due to the integration process. Another way of viewing this is that position updates are very informative for improving the inertial solution. Most commonly, such position information is provided via external sensors such as GNSS, UWB, cameras, sonar, and lasers. RF-based systems that provide absolute position updates typically require additional infrastructure that may not be available indoors. Optical systems that can provide relative position updates are more flexible but require direct line-of-sight, which means the sensor cannot be placed inside pockets or bags. Relative position updates from magnetometers provide a viable solution to all the above problems.

Magnetic signals have been used for indoor positioning by fingerprinting techniques (Zhang et al., 2016) (requires a pre-surveying phase) and SLAM techniques (does not require a pre-surveying phase) (Ferris et al., 2007). In SLAM, the pose is often estimated using a particle filter and the magnetic map is modelled separately (Vallivaara et al., 2011). The method presented in this paper combines the pose estimation with the magnetic map modelling in the Gaussian Process Regression (GPR) framework. Compared to popular LiDAR-based SLAM workflows it has a higher computation load, but has the advantage that many of the key steps are integrated into a single least-squares adjustment to allow tuning parameters to adapt to the dataset automatically. For example, most of the manually tuned parameters, such as neighbourhood size for surface modelling, are trained by the data.

## 2. MATHEMATICAL MODEL

### 2.1 Inertial Data

Most modern day IMUs perform strap-down integration at a high frequency (e.g. kilohertz) and then report odometry information as change in rotation ($dq$) and change in velocity ($dv$) at a lower, user-desired rate. Given the initial conditions, the orientation ($q$), velocity ($v$), and position ($p$) of an IMU can be computed efficiently using Equation 1. To compensate for the biases in the accelerometers ($b_a$) and gyroscopes ($b_\omega$), a random walk model is adopted (Equation 2). Note: $T$ is the strap-down integration period, $g$ is the gravity vector, $R$ is the rotation matrix, and the superscripts $L$ and $S$ represent the local frame and sensor frame, respectively.

$$\begin{aligned} {}^{LS}q_{t+T} &= {}^{LS}q_t \, dq_t \\ {}^{L}v_{t+T} &= {}^{L}v_t + {}^{LS}R_t(dv_t) + T\,{}^{L}g \\ {}^{L}p_{t+T} &= {}^{L}p_t + T\,{}^{L}v_t + {}^{LS}R_t\left(\frac{T}{2}dv_t\right) + \frac{T^2}{2}\,{}^{L}g \end{aligned} \quad (1)$$

$$^sb_{\omega,t+T} = {^sb_{\omega,t}} + w_\omega$$
$$^sb_{a,t+T} = {^sb_{a,t}} + w_a \qquad (2)$$

## 2.2 Magnetic Data

A map of the ambient magnetic field strength can be easily constructed using spatial interpolation techniques (e.g. linear or bi-cubic interpolation) when the trajectory of the sensor is known (e.g. measured by a camera-based position tracking system (Solin et al., 2015)). However, such absolute position information is typically expensive or difficult to obtain. It is more common to have relative position information in indoor navigation.

Assuming a calibrated tri-axial magnetometer that is time synchronized with the IMU is available, the relationship between the observed magnetic field vector in sensor frame ($^sy_m$) and the magnetic field in the mapping frame ($^Lm$) can be expressed using Equation 3, where $x$ represents the location at which the magnetic field was sampled and the measurement noise $\varepsilon_m$ is assumed to follow a Gaussian distribution. It is worth mentioning that any small inertial errors may accumulate over time, resulting in significant drifts in the navigation solution.

$$^sy_m = {^{sL}R}\, {^l m(x)} + \varepsilon_m \qquad (3)$$

Typically in machine learning textbooks the GPR solution is derived following the Bayesian inference principles (Rasmussen and Williams, 2006). Beginning with Bayes' rule (Equation 4), the marginal likelihood is integrated over the unknown parameters. The best estimate of the unknowns is then determined by maximizing Equation 5.

$$f(\theta|x,y) = \frac{f(y|\theta,x)f(\theta|x)}{f(y|x)} \qquad (4)$$

$$f(y|x) = \int f(y|\theta,x)f(\theta|x)d\theta$$
$$\hat{\theta} = \arg\max y^T(K+C_l)^{-1}y - \log|K+C_l| \qquad (5)$$

Although the computational complexity of this is O(n³) and the memory requirement is O(n²); this is an efficient GPR formulation when the objective is to predict the magnetic field strength at a new location given all the training magnetic measurements at known locations. In the case of a tightly-coupled IMU and magnetometer SLAM solution, it is more beneficial to follow the frequentist's interpretation, where the probability of the prior multiplied by the likelihood is maximized (Equation 6); this is the maximum a posteriori estimate of the trajectory and the magnetic field map. Note: $r$ is the residuals, $C_l$ is the covariance matrix of the measurements, and the unknowns are $\theta = [{^Lp}, {^Lv}, {^{LS}q}, {^sb_\omega}, {^sb_a}, {^Lm}, \sigma_f, l]$

$$f(\theta|x,y) \propto f(y|\theta,x)f(\theta|x)$$
$$\hat{\theta}_{MAP} = \arg\max f(\theta|x,y) = \arg\min \hat{r}^T C_l^{-1}\hat{r} + {^Lm^T}K^{-1}m + \log|K| \qquad (6)$$

The ambient magnetic fields are assumed to be realizations of a Gaussian random process prior with a given covariance matrix (i.e. kernel) K, where K is typically chosen to be the squared exponential kernel (Equation 7). This enforces simple smoothness in the local magnetic field.

$$K(x,x') = \sigma_f^2 e^{\frac{-\|x-x'\|^2}{2l^2}} \qquad (7)$$

Even though this is a valid assumption, recent research has shown that for magnetic fields the three directional components are actually correlated. Based on Maxwell's equation the magnetic field can be split into the B field and the H field; the B field needs to be divergence-free (no sinks) and the H field needs to be curl-free (no swirls) (Wahlström et al., 2013). The curl-free and divergence-free kernels are given in Equation 8. Instead of treating the x, y, and z magnetic fields as three independent smooth surfaces, their physical correlation is included to provide a more accurate model of the local magnetic map, which translates into a more accurate trajectory estimate.

$$K_H(x,x') = \frac{1}{l^2}\left(I_{n_x} - \left(\frac{x-x'}{l}\right)\left(\frac{x-x'}{l}\right)^T\right)\sigma_f^2 e^{-\frac{1}{2l^2}\|x-x'\|^2}$$
$$K_B(x,x') = \frac{1}{l^2}\left(\left(n_x - 1 - \left\|\frac{x-x'}{l}\right\|^2\right)I_{n_x} + \left(\frac{x-x'}{l}\right)\left(\frac{x-x'}{l}\right)^T\right)\sigma_f^2 e^{-\frac{1}{2l^2}\|x-x'\|^2} \qquad (8)$$

## 3. RESULTS AND ANALYSIS

### 3.1 Simulated Datasets

In GP-SLAM the hyper-parameters characterize the ambient magnetic field and are estimated using the data rather than being tuned manually. It can be valuable to understand the impacts of different magnetic fields on the SLAM solution through simulation to better understand the requirements for the proposed methodology. In all simulations below, the sensor is assumed to be exhibiting translational motion only, in a plane where the magnetic field is measured every 25cm, the odometry noise is set to be 0.5mm (0.2% of the distance travelled) with a bias of 5.0mm in both X and Y directions, $\sigma_f$ is 0.1, and $l$ is 10cm, unless otherwise stated.

**3.1.1 Effects of varying $\sigma_f$:** This parameter describes the amplitude of the magnetic signal. For instance, $\sigma_f$ is expected to be small when the magnetic field is homogeneous (e.g. outdoors) and large when near ferromagnetic materials (e.g. computers and motors). $\sigma_f$ can also be interpreted as the noise of the Gaussian Process constraint: the higher the value, the lower the weight of the spatial correlation in the least-squares adjustment, and vice versa.

Four different cases of $\sigma_f$ are shown below where GP-SLAM is performed with the hyper-parameters being fixed to their true value. From Figure 1 and Table 1 it can be observed that the variation of the ambient field must be sufficiently large relative to the sensor noise, but beyond a certain threshold, having more magnetic disturbances does not improve the accuracy of the SLAM solution.

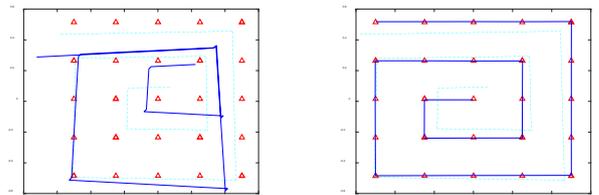

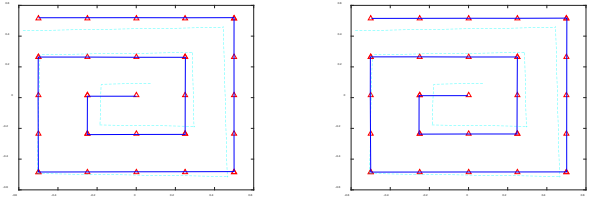

Figure 1: Estimated trajectory using GP-SLAM with different true $\sigma_f$. (Top Left) $\sigma_f = 0.001$, (Top Right) $\sigma_f = 0.100$, (Bottom Left) $\sigma_f = 1.000$, and (Bottom Right) $\sigma_f = 10.00$. The cyan dash line shows the odometry only solution, the blue solid line indicates the GP-SLAM solution, and the red triangles are the ground truth.

|  | RMSE [m] | Max Error [m] |
|---|---|---|
| Before SLAM | 0.066 | 0.112 |
| After SLAM, $\sigma_f = 0.001$ | 0.189 | 0.318 |
| After SLAM, $\sigma_f = 0.100$ | 0.003 | 0.005 |
| After SLAM, $\sigma_f = 1.000$ | 0.004 | 0.006 |
| After SLAM, $\sigma_f = 10.00$ | 0.003 | 0.005 |

Table 1: Error in estimated trajectory for various $\sigma_f$

**3.1.2 Effects of a wrong $\sigma_f$:** The quality of the estimated hyper-parameters has a direct impact on the overall GP-SLAM solution. In this simulation, the true $\sigma_f$ is 0.1, but the $\sigma_f$ in the GP-SLAM is fixed to four different values as shown below. It appears from the results that it is better to overestimate $\sigma_f$ rather than underestimating; this is likely because it is better to be conservative and place less weight on the magnetic field position updates rather than "over-trusting" the magnetic field.

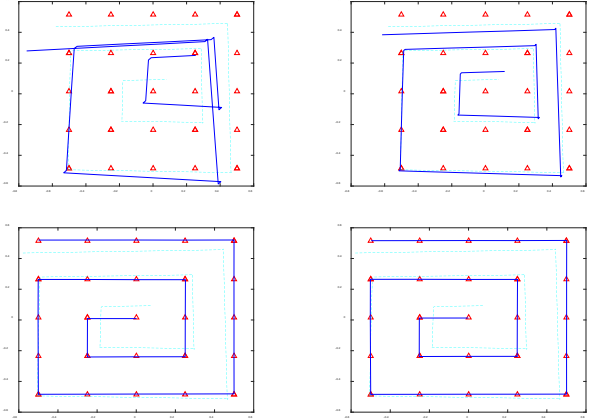

Figure 2: Estimated trajectory using GP-SLAM with the wrong $\sigma_f$. (Top Left) $\sigma_f = 0.001$, (Top Right) $\sigma_f = 0.010$, (Bottom Left) $\sigma_f = 1.000$, and (Bottom Right) $\sigma_f = 10.00$. The cyan dash line shows the odometry only solution, the blue solid line indicates the GP-SLAM solution, and the red triangles are the ground truth.

|  | RMSE [m] | Max Error [m] |
|---|---|---|
| Before SLAM | 0.066 | 0.112 |
| After SLAM, $\sigma_f = 0.001$ | 0.206 | 0.346 |
| After SLAM, $\sigma_f = 0.010$ | 0.103 | 0.174 |
| After SLAM, $\sigma_f = 1.000$ | 0.004 | 0.006 |
| After SLAM, $\sigma_f = 10.00$ | 0.004 | 0.006 |

Table 2: Error in estimated trajectory when using the wrong $\sigma_f$

**3.1.3 Effects of varying $l$:** This length parameter determines the radius at which the magnetic field samples should be spatially correlated. A large $l$ would indicate that even far points would have an impact on the query point. A small $l$ usually occurs when the local magnetic field is varying rapidly; therefore, only nearby points are useful at predicting the magnetic field signal at the target location.

Two different cases are simulated where the $l$ parameter is fixed to the true values of 0.1m and 0.4m (Figure 3 and Table 3). Smaller $l$ means a smaller convergence region for GP-SLAM but it can yield more accurate results than a larger $l$. Furthermore, a better initial trajectory is needed to ensure convergence to the global minimum. A larger $l$ can represent a homogeneous magnetic field; therefore the accuracy of the GP-SLAM solution is compromised due to the lack of uniqueness in the magnetic signature.

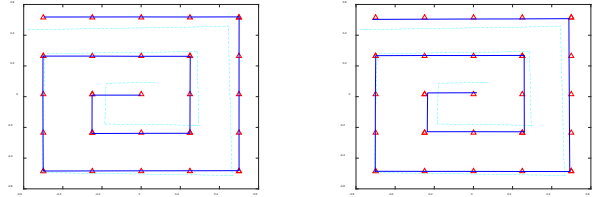

Figure 3: Estimated trajectory using GP-SLAM with different true $l$. (Left) $l = 0.1$m, and (Right) $l = 0.4$m. The cyan dash line shows the odometry only solution, the blue solid line indicates the GP-SLAM solution, and the red triangles are the ground truth.

|  | RMSE [m] | Max Error [m] |
|---|---|---|
| Before SLAM | 0.066 | 0.112 |
| After SLAM, $l = 0.1$m | 0.003 | 0.005 |
| After SLAM, $l = 0.4$m | 0.012 | 0.021 |

Table 3: Error in estimated trajectory for various $l$

**3.1.4 Effects of a wrong $l$:** Estimating the wrong length parameter is comparable to setting the wrong neighbourhood search size. In this case the magnetic field was simulated using $l = 0.1$m. In each trial, $l$ is fixed to a different (wrong) value and its effects on the global SLAM solution are reported in Figure 4 and Table 4.

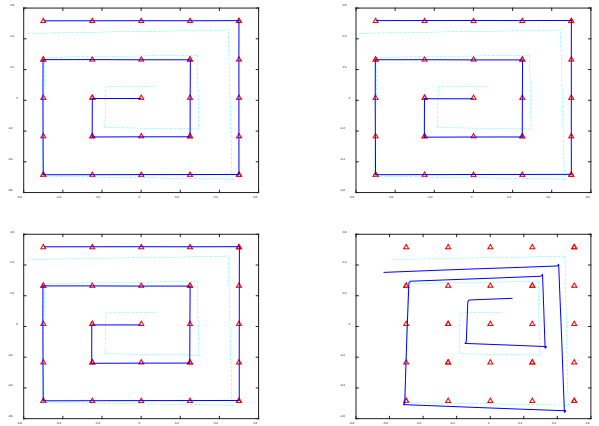

Figure 4: Estimated trajectory using GP-SLAM with the wrong $l$. (Top Left) $l = 0.005$m, (Top Right) $l = 0.050$m, (Bottom Left) $l = 0.150$m, and (Bottom Right) $l = 0.400$m. The cyan dash line shows the odometry only solution, the blue solid line indicates

the GP-SLAM solution, and the red triangles are the ground truth.

|   | RMSE [m] | Max Error [m] |
|---|---|---|
| Before SLAM | 0.066 | 0.112 |
| After SLAM, $l$ = 0.005m | 0.003 | 0.004 |
| After SLAM, $l$ = 0.050m | 0.003 | 0.005 |
| After SLAM, $l$ = 0.150m | 0.004 | 0.006 |
| After SLAM, $l$ = 0.400m | 0.127 | 0.214 |

Table 4: Error in estimated trajectory when using the wrong $l$

Small errors in $l$ do not have a significant impact on the SLAM solution. While GP-SLAM does not appear to be hypersensitive to the neighbourhood radius, it is better to underestimate $l$ than to overestimate; this is likely because it is better to ignore valuable information from nearby points rather than to allow spatially uncorrelated measurements from a larger neighbourhood to influence the query point.

**3.1.5 Effects of varying odometry noise:** Like most target-less optical SLAM solutions, GP-SLAM is sensitive to the initial pose. When the initial trajectory is significantly different from the true trajectory, GP-SLAM tends to converge to a local minimum because it uses a nonlinear least-squares formulation. Unlike frame-based optical SLAM methods though, each magnetometer reading represents the measurement of a single point, which is insufficient to estimate the 3D position and 3D orientation. Hence, it relies heavily on the odometry information to link temporally close measurements. As shown in Table 5, a smaller odometry noise means a larger convergence region (i.e. larger biases in the odometer can be handled). Although a small odometry noise is always beneficial, beyond a certain threshold, more precise odometry does not seem to improve the solution significantly.

|   |   | RMSE [m] | Max Error [m] |
|---|---|---|---|
| 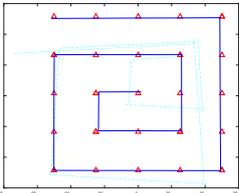 Odometry noise = 0.5mm | Before SLAM | 0.191 | 0.328 |
|  | After SLAM | 0.012 | 0.021 |
| 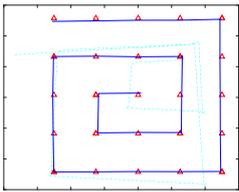 Odometry noise = 2.5mm | Before SLAM | 0.196 | 0.334 |
|  | After SLAM | 0.012 | 0.024 |
| 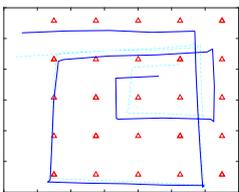 Odometry noise = 5.0mm | Before SLAM | 0.201 | 0.338 |
|  | After SLAM | 0.154 | 0.222 |

Table 5: Errors in estimated trajectory when varying the odometry noise. The bias in X and Y are set to 15mm and $l$ is 0.4m.

### 3.2 Real Datasets

In the following experiments, a MEMS-based IMU from Xsens Technologies, the MTi-300 Attitude and Heading Reference System was used.

#### 3.2.1 Indoor Environments:

**Experiment 1**

In the first test, the IMU was placed on a typical office desk with computer and smartphone within its vicinity. It was then moved along a rectangular trajectory four times with complete overlap between each loop. Even from looking at the magnitude of the magnetic measurements it is possible to see the similarities between the four loops (Figure 5). In this trivial example, it is possible to perform loop-closure detection using auto-correlation since the variation in the magnetic field is as high as 1.67 a.u. (Jung and Myung, 2015).

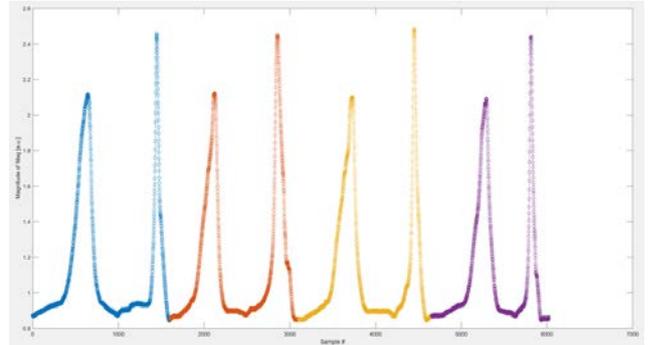

Figure 5: Magnitude of the measured magnetic signal when looping the same path four times.

To obtain a sensible initial trajectory, a weak zero position update was applied to constrain the IMU position to be within a one metre radius sphere. The IMU trajectory before and after GP-SLAM is shown in Figure 6. Both the horizontal and vertical accuracies were improved from the centimetre-level to millimetre-level after 1800 iterations. The high number of iterations is an undesirable trait of GP-SLAM as the Lagrangian cost reduction is quite slow after every iteration (Figure 7).

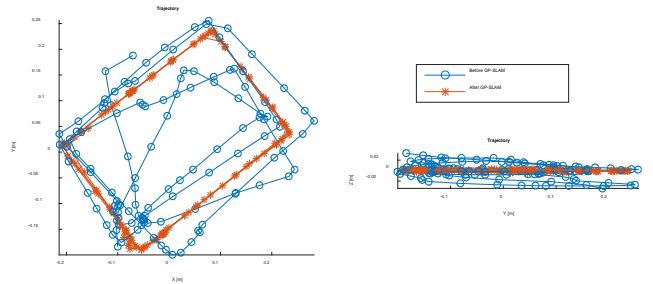

Figure 6: Estimated trajectory from repeating a rectangular path four times in a heterogeneous magnetic environment.

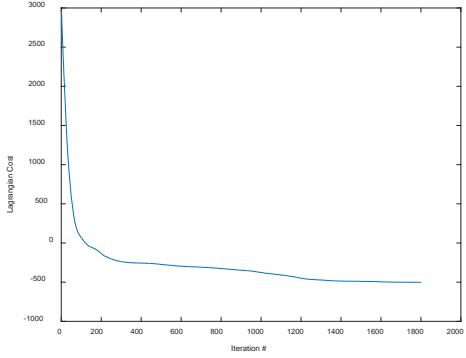

Figure 7: The optimization cost function at every iteration

**Experiment 2**

In the second experiment, the IMU was moved around in a spiral pattern similar to the simulated trajectory. This test highlights one of the strengths of GP-SLAM, which is the unnecessity for perfect overlap between each pass of the same area. Unlike the first experiment, by studying the magnetic signal alone it is difficult to perform loop-closure detection (Figure 8). A threshold for the auto-correlation will need to be set and the possibility of revisiting the same position in a different direction will need to be handled in the logic (Jung and Myung, 2015). The estimated trajectory before and after using the magnetic field to perform GP-SLAM is provided in Figure 9. The adjustment took 797 iterations to converge and the RMSE of the trajectory is reduced from the centimetre-level to millimetre-level. One of the main factors that contributed to the difference in trajectory is the recovered accelerometer and gyroscope biases (Table 6).

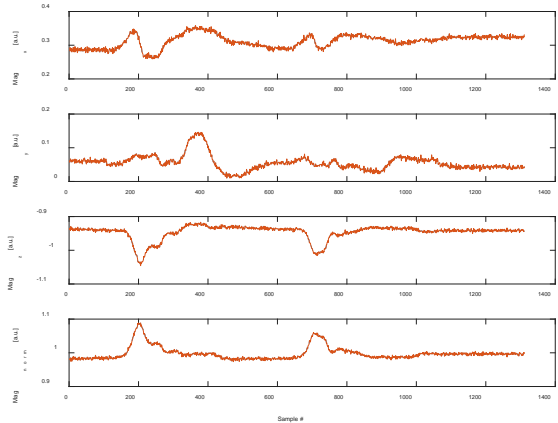

Figure 8: The measured magnetic signal when moving the IMU in a spiral pattern under a heterogeneous magnetic field.

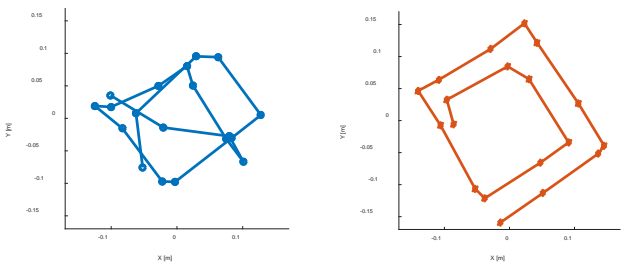

Figure 9: The reconstructed spiral trajectory. (Left) before GP-SLAM in blue and (Right) after GP-SLAM in orange.

|   | **Before GP-SLAM** | | **After GP-SLAM** | |
| --- | --- | --- | --- | --- |
|   | Gyr Bias [deg] | Acc Bias [m/s$^2$] | Gyr Bias [deg] | Acc Bias [m/s$^2$] |
| X | 0.0135 | -0.0030 | -0.0509 | 0.1043 |
| Y | 0.0959 | 0.0020 | 0.0023 | 0.0709 |
| Z | -0.0247 | 0.0255 | 0.1326 | 0.0255 |

Table 6: The recovered gyroscope and accelerometer biases in the adjustment

**Experiment 3**

The third experiment was performed on a special aluminum table that is free of ferromagnetic materials. In addition, no magnetic objects were placed within one metre of the table. The IMU was moved in a square pattern multiple times. When studying the magnitude of the magnetic measurements the field appears to be fairly homogeneous. However, when examining the signal in detail, a range of 0.07 a.u. between the maximum and minimum magnetic reading can be perceived. Such low magnetic disturbance poses a challenging situation for GP-SLAM. In all other test cases, it did not matter if the squared exponential kernel or curl-free and divergence-free kernels were applied, the differences in the trajectories were insignificant. However, in a near-homogeneous field, a more noticeable impact of the chosen kernel can be perceived. Figure 10 shows the GP-SLAM solution using the two different kernels. In this scenario neither kernel was able to reconstruct a perfect square-shaped trajectory. The squared exponential kernel resulted in a higher maximum error but preserved the orthogonality of all edges, while the curl-free and divergence-free kernel closed all the loops properly but deformed the overall geometry of the trajectory into an arbitrary quadrilateral. The latter solution might be preferred if it is to be combined with other optical SLAM solutions, because it is more consistent internally (i.e. all loops were properly detected).

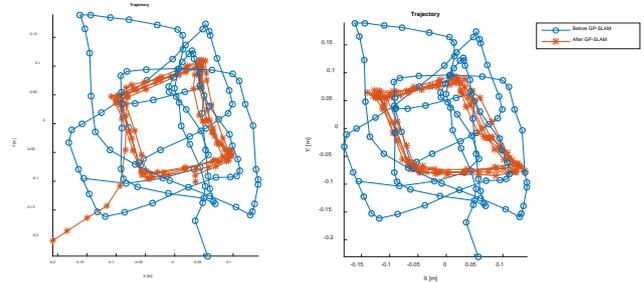

Figure 10: The recovered square trajectory under a near homogeneous magnetic field using GP-SLAM. (Left) Using the squared exponential kernel and (Right) curl-free and divergence-free kernel.

**Experiment 4**

The final indoor experiment took place on a typical wooden office desk. The MTi-300 was treated like a pen, as it was moved along the surface writing the word "Xsens". After GP-SLAM the handwritten characters became legible even though no active beacons were introduced.

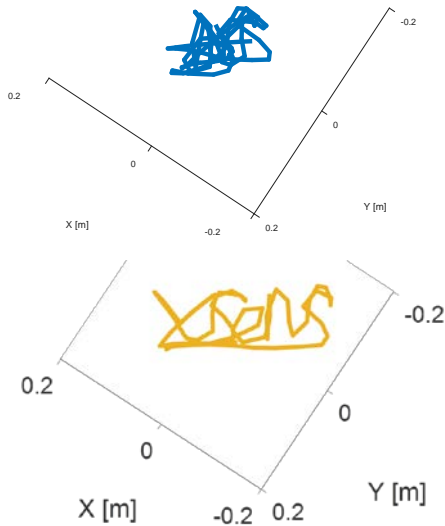

Figure 11: Recovered complex handwriting pattern using the ambient magnetic field. (Top) before performing GP-SLAM and (Bottom) after GP-SLAM.

### 3.2.2 Outdoor Environments:

In a real homogeneous magnetic field environment, GP-SLAM has a favourable property of not making the initial solution any worse. To demonstrate this, the aluminium table was moved outdoors into an open field where the IMU was moved along a rectangular track several times. In this case, the $l$ of the hyper-parameter approaches a large number rapidly, $\sigma_f$ approaches zero, and the adjustment converges. Mathematically it can be seen from Equation 7 that as $l$ approaches infinity all the covariances approach $\sigma_f$ (which is zero), effectively disabling all spatial correlations automatically. In conventional loop-closures in SLAM, a wrong association can have a devastating effect on the entire solution due to the strong assumption being made in the global relaxation step. In GP-SLAM, the assumption being made is much weaker and therefore has a smaller impact on the solution when the assumptions are violated; i.e. it is more robust.

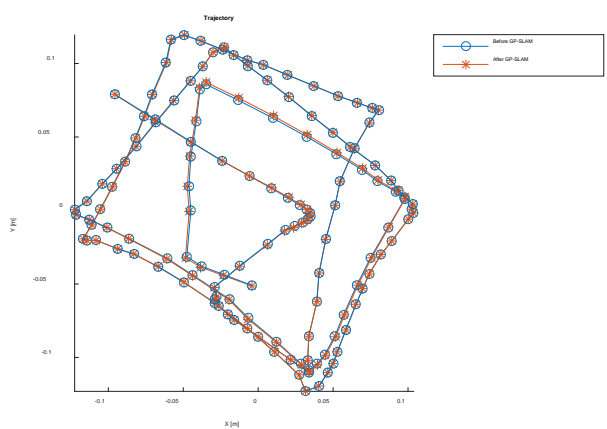

Figure 12: Estimated rectangular trajectory pattern in a homogeneous magnetic field environment

## 4. CONCLUSION AND FUTURE WORK

This paper addresses the problem of indoor navigation by exploiting magnetic disturbances in a GP-SLAM framework with a tightly-coupled MEMS IMU. The orientation, velocity, position, sensor biases, hyper-parameters, and magnetic field map were all estimated simultaneously while performing continuous loop-detection and loop-closure in a batch least-squares adjustment. Through different experiments it was shown that magnetic field SLAM in a GPR framework can improve the overall trajectory without making the solution significantly worse. Although it has no effect on the overall trajectory when applied in typical outdoor environments, it can be argued that satellite signals are available outdoors and the magnetometer can be used for heading updates instead.

Future work will focus on improving the efficiency of GP-SLAM through sparse kernel approximations in order to scale it to larger problems. The current solution assumes the magnetic field is not changing; to make it applicable to more consumer applications (e.g. wearable technologies) the static magnetic map assumption will be lifted.

### ACKNOWLEDGEMENTS

This research is funded by TRAcking in complex sensor systems (TRAX), under the EU's Seventh Framework Programme (grant agreement No. 607400), and the Natural Science and Engineering Research Council (NSERC) of Canada. Valuable discussions with Jeroen Hol and Henk Luinge are gratefully acknowledged.